\documentclass[letterpaper, 10 pt, conference]{ieeeconf} \IEEEoverridecommandlockouts
\usepackage{cite}
\usepackage{amsmath,amssymb,amsfonts}
\usepackage{algorithmic}
\usepackage{graphicx}
\usepackage{textcomp}
\usepackage{xcolor}
\usepackage{url}
\usepackage{booktabs}
\usepackage{multirow}
\usepackage{adjustbox}
\usepackage{siunitx}

\usepackage{amsmath}
\usepackage{amssymb}
\usepackage{tensor}

\DeclareMathAlphabet{\mathcal}{OMS}{cmsy}{m}{n}

\def\V{\ensuremath{\mathcal{V}}}

\def\cI{\ensuremath{\mathcal{I}}}

\newcommand{\mat}[1]{\ensuremath{\mathbf{#1}}}

\newcommand{\set}[1]{\ensuremath{\left\{#1\right\}}}

\newcommand{\tf}[3]{\tensor[^{#1}]{\mat{#2}}{_{#3}}}

\newcommand{\tuple}[1]{\left\langle#1\right\rangle}

\DeclareMathOperator*{\argmax}{arg\,max}

\usepackage{xspace}

\newcommand{\ie}{\mbox{i.\,e.}\xspace}

\renewcommand{\[}{\begin{equation}}
\renewcommand{\]}{\end{equation}}

\renewcommand{\baselinestretch}{0.99}

\usepackage[capitalize]{cleveref}

\crefname{figure}{Fig.}{Figs.}
\Crefname{figure}{Figure}{Figures}
\crefname{section}{Sec.}{Secs.}
\Crefname{section}{Section}{Sections}
\Crefname{table}{Table}{Tables}
\crefname{algorithm}{Algo.}{Algos.}
\Crefname{algorithm}{Algorithm}{Algorithms}
\crefname{appendix}{Sec.}{Secs.}
\Crefname{appendix}{Section}{Sections}

\title{\LARGE \bf
Preference-Based Long-Horizon Robotic Stacking with Multimodal Large Language Models
}

\author{Wanming Yu$^{1}$, Adrian Röfer$^{2}$, Abhinav Valada$^{2}$ and Sethu Vijayakumar$^{1}$
\thanks{*This work was supported by JST Moonshot R\&D (Grant No. JPMJMS2031) and by the BrainLinks-BrainTools center of the University of Freiburg.}
\thanks{$^{1}$Wanming Yu and Sethu Vijayakumar are with School of Informatics, University of Edinburgh, UK
        {\tt\small \{wanming.yu, sethu.vijayakumar\}@ed.ac.uk}}%
\thanks{$^{2}$Adrian Röfer and Abhinav Valada are with the Robot Learning Lab, University of Freiburg, Germany
        {\tt\small \{aroefer, valada\}@cs.uni-freiburg.de}}%
}

\begin{document}

\maketitle
\thispagestyle{empty}
\pagestyle{empty}

\begin{abstract}
Pretrained large language models (LLMs) can work as high-level robotic planners by reasoning over abstract task descriptions and natural language instructions, etc. However, they have shown a lack of knowledge and effectiveness in planning long-horizon robotic manipulation tasks where the physical properties of the objects are essential. An example is the stacking of containers with hidden objects inside, which involves reasoning over hidden physics properties such as weight and stability. To this end, this paper proposes to use multimodal LLMs as high-level planners for such long-horizon robotic stacking tasks. The LLM takes multimodal inputs for each object to stack and infers the current best stacking sequence by reasoning over stacking preferences. Furthermore, in order to enable the LLM to reason over multiple preferences at the same time without giving explicit instructions, we propose to create a custom dataset considering stacking preferences including weight, stability, size, and footprint, to fine-tune the LLM. 
Compared to the pretrained LLM with prompt tuning, we demonstrate the improved stacking completion of the LLM fine-tuned with our custom dataset via large-scale simulation evaluation. Furthermore, we showcase the effectiveness of the proposed framework for the long-horizon stacking task on a real humanoid robot in an online manner. 
\end{abstract}

\section{Introduction}
Stacking of boxes or containers is a common problem in various domains, ranging from private homes to shops to warehouses. The challenge in this problem lies with a multitude of factors to consider, such as the size, weight, and structural stability of the items being stacked.
Moreover, the properties of the objects being stacked might not even be readily apparent from visual observation, as all moving boxes look the same, but when lifted, one notices that some contain books and others bedding. Despite all these challenges and considerations, humans can easily trade off structural constraints with preferences and successfully reorganize a storage room. 

Planning approaches have been proposed for performing robotic stacking tasks, such as symbolic planning \cite{fikes1971strips,jiang2019task}, task and motion planning \cite{toussaint2015logic}. However, these planning approaches usually require expert knowledge or engineering efforts, suffer from limited flexibility or scalability to plan more complex scenarios on the fly. Another line of related work is using reinforcement learning which usually brings more robustness and flexibility, but can be difficult to scale to long-horizon task learning or scenarios with more objects.

The recent rise of large-scale models such as Large Language Models (LLMs) has demonstrated promising reasoning capabilities about common-sense knowledge~\cite{huang2022towards, zawalski2024robotic,honerkamp2024language,chisari2025robotic}. 
Vision Language Models (VLMs) \cite{shi2025hi} combine LLMs with vision encoders to aid the reasoning capabilities with visual inputs. VLMs have been shown to work effectively as high-level planners for generating subgoals for certain tasks, but lack knowledge of physical understandings. 
Built on top of VLMs, Vision Language Action Models (VLAs) are mainly developed aiming at inferring actions for general robotic tasks, taking physics into consideration \cite{kim2024openvla,team2025gemini,black2410pi0}. However, they require a huge amount of data to train and have to fine-tune for a specific task in order to increase the success rate and improve the performance.

Moreover, the objects in these previous works usually have distinct visual appearances, such as distinct shapes and colors. On the other hand, in the case of stacking containers, objects hidden inside a container can affect the final stacking order, with the latent characteristics of these objects only becoming evident during manipulation. Previous approaches have rarely focused on planning or reasoning based on physics properties of objects via multimodal sensing \cite{nazarczuk2024closed}. Therefore, long-horizon stacking of objects with hidden properties remains an understudied problem in robotics research. 

\begin{figure}
    \centering
    \includegraphics[width=\linewidth]{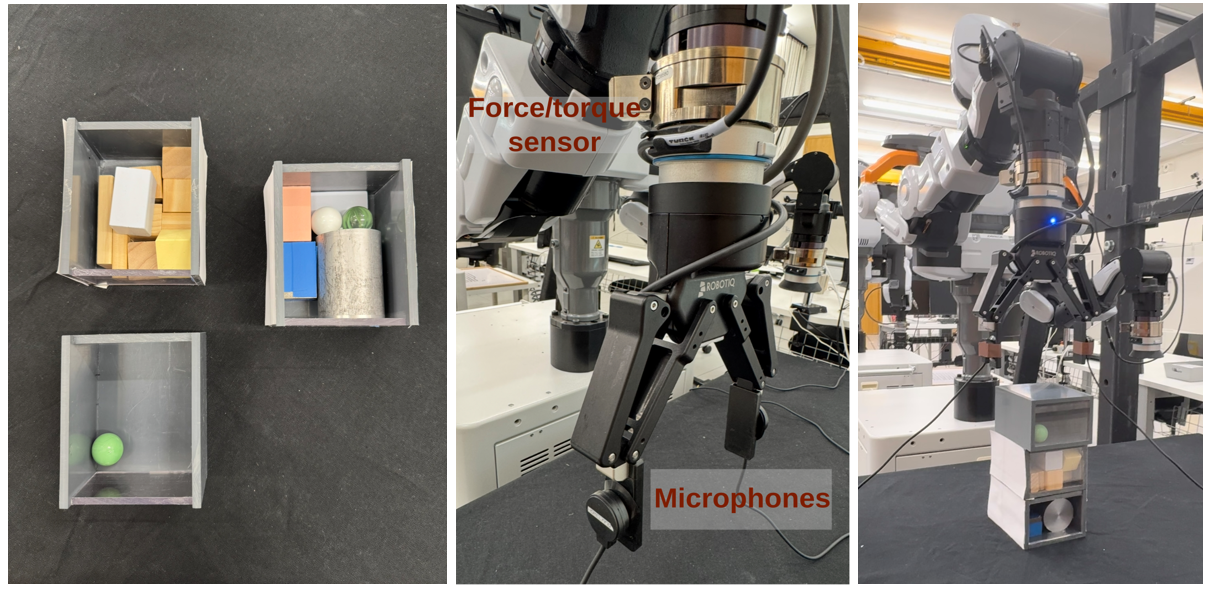}
    \caption{Long-horizon stacking task by multimodal large language models. \textbf{Left:} Three containers to be stacked with various contents. \textbf{Middle:} Multimodal sensing setup for object hidden properties. \textbf{Right:} The NEXTAGE robot performing box stacking guided by an LLM.}
    \label{fig:teaser}
\end{figure}

In studying this problem in more detail, we propose a multimodal large language model as the orchestrator of the high-level behaviors for performing long-horizon robotic stacking of objects with latent properties. We obtain these properties using force/torque sensors and microphones to get the weight and stability of each container, respectively. The multimodal system we present, displayed in \cref{fig:teaser}, is able to infer the current best stacking sequence on-the-fly according to the hidden properties, human preferences, and the current stacking status.
Moreover, we introduce a lightweight paradigm to generate additional training data in physics simulation for the model to become more accustomed to considering physical stability and plausibility of the generated plans. In a simulation environment, we study the complexity of the task, understand the challenges for off-the-shelf models, and gauge the impact of fine-tuning these models using a dataset generated from simulations. Furthermore, we showcased successful long-horizon stacking on a real humanoid robot. Our contributions are as follows:

\begin{itemize}
    \item An approach for obtaining and encoding multimodal information from audio measurements and force sensing, to enable planning on such information with a large language model (LLM) to satisfy natural language goals in a long-horizon stacking task.
    \item An efficient dataset generation approach based on physical simulation and remixing of simulation samples, to generate sufficient data for fine-tuning an LLM agent for utilizing multimodal information.
    \item Evaluation in simulation and on a real robotic system showcasing improved reasoning capabilities of fine-tuned LLM for stacking with latent properties according to natural language preferences.
\end{itemize}

\section{Related Work}
\subsection{Planning Approaches for Robotic Stacking}

Symbolic planners, such as STRIPS \cite{fikes1971strips} and PDDL-based planners \cite{jiang2019task}, have been used to solve long-horizon robotic tasks by searching over discrete action sequences. However, they require careful symbol grounding to map symbols to real-world actions and require backtracking if the symbolic plan is geometrically infeasible.
A causal probabilistic framework has been proposed to extend classical symbolic planning by explicitly modeling causality and uncertainty and has also been used to solve block stacking tasks \cite{cannizzaro2023towards,cannizzaro2024causal}. A similar work to ours proposed a closed-loop neuro-symbolic interactive reasoning framework for task planning, taking object weight into account \cite{nazarczuk2024closed}. However, only pick-and-place tasks are demonstrated, and no overall stability or other preferences are considered.

Task and motion planning combine discrete symbolic decisions with continuous geometric constraints. For instance, logic-geometric programming \cite{toussaint2015logic} demonstrated sequential robot manipulation such as creating a physically stable construction from assorted blocks to maximize height. However, it is an offline approach and has difficulty scaling to longer action sequence horizons. In general, planning approaches require domain knowledge and engineering efforts, and can be difficult to scale as the number of objects or task horizon increases.

\subsection{Reinforcement Learning for Robotic Stacking}
Reinforcement learning has been used for RGB stacking, where geometric shapes are considered during robotic stacking. In \cite{lee2021beyond}, a robot manipulation benchmark was introduced for completing the stacking of objects with diverse shapes by a vision-based RL agent. In \cite{lampe2024mastering}, iterative reinforcement learning was proposed to learn to stack objects of diverse shapes in an end-to-end manner directly from pixels on real robots.
However, these are based on the visual properties of the objects and do not consider the case where boxes may have the same appearance but different content inside. Moreover, reinforcement learning approaches suffer from scaling to long-horizon tasks.
\subsection{Vision Language Action (VLAs) Models for General Robotic Tasks}
Vision Language Action models are usually trained based on pretrained VLMs. Since VLMs do not include knowledge about robotics, a significant amount of data from various robotic tasks of different types of robots needs to be collected on top of the VLM data in order to handle general robotic tasks \cite{kim2024openvla,black2410pi0,team2025gemini,argus2025cvla}. VLAs are difficult to deploy zero-shot on new tasks or on new robots, and typically need further data for fine-tuning the model. Furthermore, these VLAs mainly focus on the tasks where relevant objects have identifiable visual differences. In our object stacking task, some of the objects may have exactly the same visual appearance but different hidden properties. 

\subsection{Robotic Stacking Applications}
Robotic stacking is a foundational and essential skill for assembly and disassembly \cite{chiang2018design}. Moreover, stacking stones of irregular shapes is also an active research field. In \cite{liu2023stability}, stability estimates were incorporated into each step of the sequential planning process for stacking stones into a specific target structure. In \cite{furrer2017autonomous}, a pose searching algorithm was introduced to propose the next stable stack of stones, and a robotic arm was used to build vertical stone towers iteratively.
Neither of these works considers human preferences expressed in natural language as a constraint for the execution of the task or to define the target structure.

\section{Problem Formulation}

\begin{figure}
    \centering
    \includegraphics[width=0.8\linewidth]{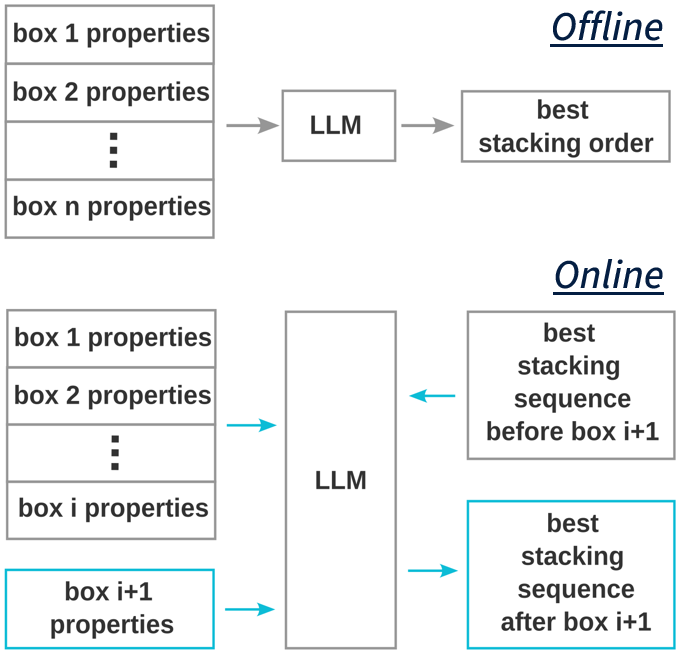}
    \caption{The LLM can plan long-horizon stacking tasks in either an offline (top) or online manner (bottom).}
    \label{fig:planner}
\end{figure}

We consider a box stacking scenario in which the robot is required to stack $B$ boxes, each with hidden objects inside, into a single stack. Each box $b$ has a visual property $\V_b$, an initial pose $\tf{W}{T}{b,0}$, and an initially unknown latent property $\cI_b$ associated with it. 
Given a user prompt $Q$ referencing multiple modalities, the aim is to generate a stacking order $S = \tuple{b_i,\ldots, b_j}$.
As the latent properties $\cI_b$ are initially unknown, the agent needs to interact with each box to measure the latent characteristics. The task can be addressed in two modes which are illustrated in \cref{fig:planner}.
\begin{itemize}
    \item \textit{Offline planning:} The agent is given the observed inertial properties for all boxes, all at once, and generates the best stacking plan.
    \item \textit{Online planning:} After the robot interacts with box $b$, the agent is given the observed inertial properties $\cI_b$ of box $b$. The agent can then update the previous stacking plan if necessary, according to available information, or simply continue to interact with another box.
\end{itemize}

\section{Methodology}

Our approach consists of a framework for obtaining and encoding multimodal information for an LLM agent and enabling this agent to control a physical robot, and a method for generating data to train this framework for our stacking task using a physics simulation. In the following sections, we first present the framework, then the data generation method.

\subsection{Framework Overview}

\begin{figure}
    \centering
    \includegraphics[width=\linewidth]{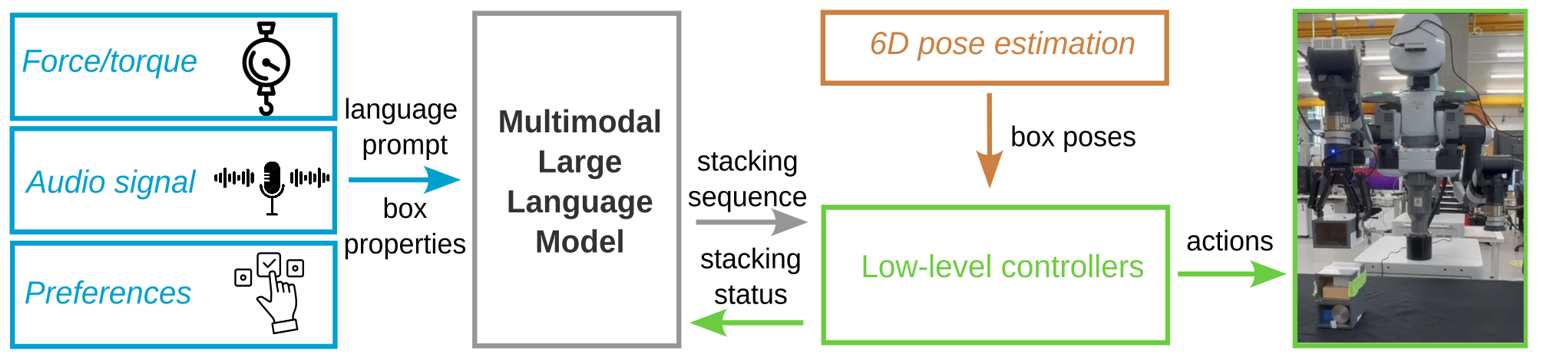}
    \caption{Overview of the proposed long-horizon robotic stacking framework.}
    \label{fig:method}
\end{figure}

The main research question we aim to address in this work is whether we can leverage large-scale foundation models to reason and plan a stacking sequence of containers according to human-specified preferences. These preferences can refer to apparent characteristics, such as size, as well as hidden physical properties such as weight and stability of a container's contents. To this end, we propose to use a large language model to consider multi-modal sensing data for planning the long-horizon object stacking tasks according to various stacking preferences. An overview of our proposed multimodal long-horizon stacking planning framework is shown in Fig. \ref{fig:method}. The LLM takes multimodal inputs for each box, generates the current best stacking sequence according to specified preferences, which is converted into motion goals by the use of object pose estimation to perform the sequence.

\subsection{Hidden Property Sensing}
The hidden properties essential to the object stacking task mainly include the object weight and stability due to unknown items inside. Weight is measured with a force-torque sensor on the wrist as the robot lifts up a box. In the offline planning case, the robot does so once for every box on the table before planning.
Stability is measured by having the robot tilt the box within $\ang{90}$ and recording the audio signals from microphones on the gripper. The audio signals are accumulated during the tilting motion. Greater movement, \ie lower stability, inside the container will generate more noise. This property is normalized to the range $[0, 1]$ where $1$ represents no motion inside the container.
Both weight and stability values are encoded to construct the language prompt.
Finally, the constructed language prompt is passed to the LLM together with the current stacking state, which is expressed as a list of box names starting with the box at the bottom and ending with the one on top.

\subsection{Prompt Definition for Long-Horizon Stacking}

\begin{figure}
    \centering
    \includegraphics[width=\linewidth]{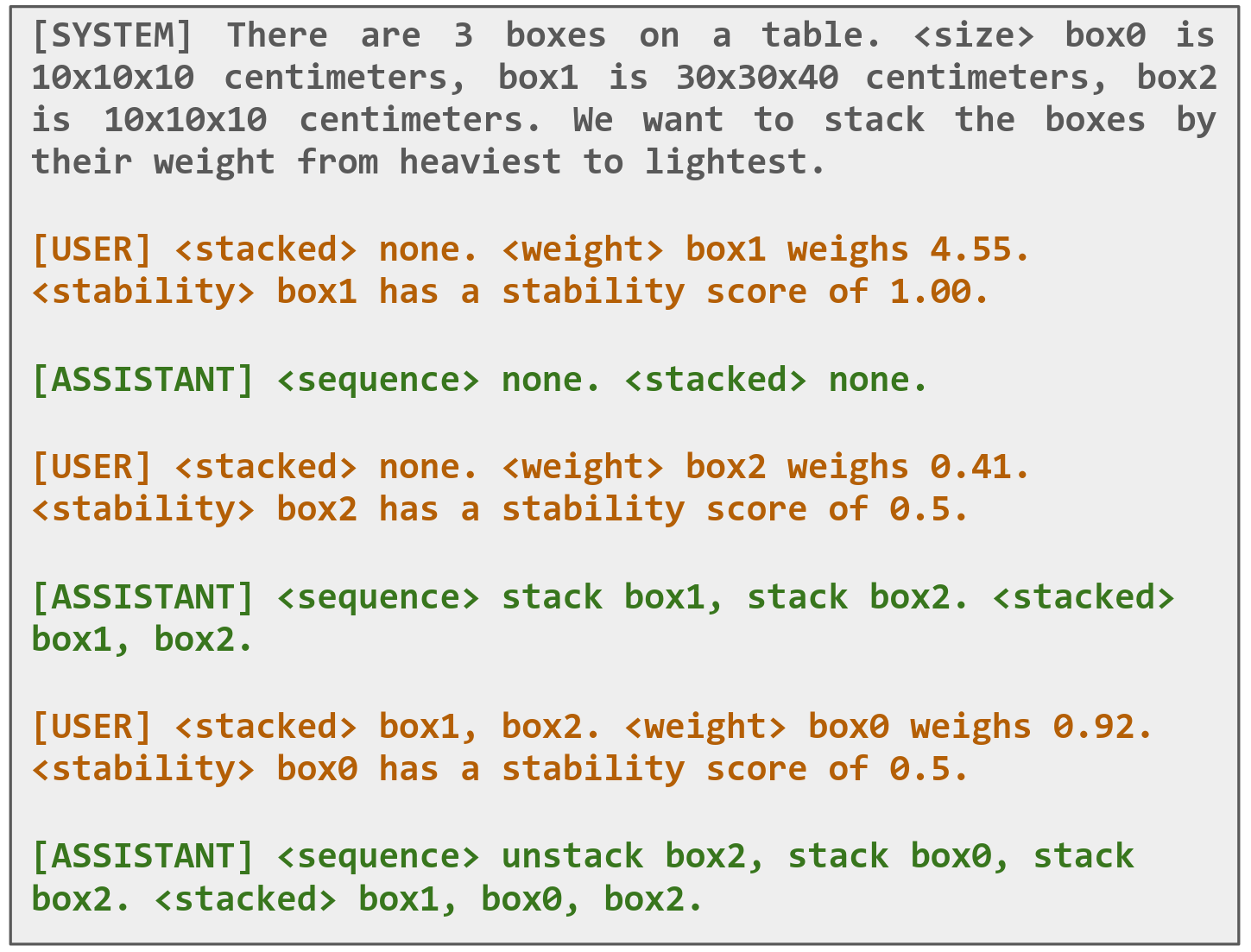}
    \caption{Example of an iterative LLM reasoning process for stacking boxes by weight, where [USER] provides box attributes and [ASSISTANT] outputs the stacking sequence to achieve the desired order.}
    \label{fig:prompt}
\end{figure}

Given object weight, object stability, and the current stacking status, the large language model will be able to infer the current best stacking sequence according to specified stacking preferences. An example of the constructed prompt is shown in Fig. \ref{fig:prompt}.

\subsection{Dataset Generation with Preferences for LLM Fine-tuning}
\label{sec:gen-data}
To obtain training data for fine-tuning the LLM, we use a physics simulation environment based on PyBullet.
In our simulation, we generate closed boxes of different sizes and varying material densities (cardboard, wood, and plastic), and fill these with smaller objects of different simple geometric shapes (spheres, boxes, and cylinders). We assign these objects a stability score
\[
    s_o=\frac{\min(w, d)}{h},
\]
where $w, d$ are the width and depth of the object's footprint and $h$ is its height. In the case of a cylinder $w, d$ equal the cylinder's diameter. If $o$ is a sphere, $s_o = 0$.
We compute the overall stability score for a box as the mean of the stability scores of its contents.

\begin{figure}
    \centering
    \includegraphics[width=0.6\linewidth]{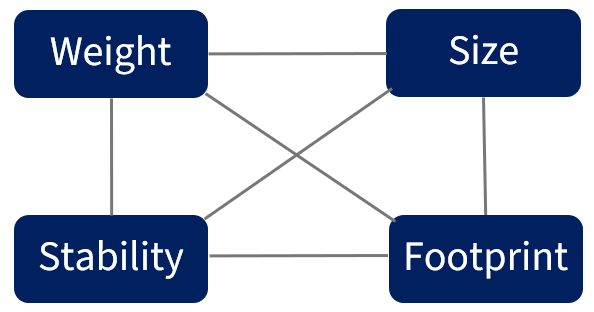}
    \caption{Stacking preferences used for generating our custom dataset.}
    \label{fig:preference}
\end{figure}

While satisfying an individual preference such as ``\textit{Stack the boxes heaviest to lightest}'' defines an ordering of boxes, it is not said that this order actually yields a stable stack. It might be that the heaviest box is a lot smaller than the other ones, which would lead to an overall unstable tower. In addition, there might also be multiple valid orders to satisfy a preference or a combination of them.
These considerations make it very difficult to directly generate desirable executions for the agents to learn from.
Instead, we simulate all possible stacks for a given initial sample of boxes and compute their scores with respect to individual or combinations of preferences (Fig. \ref{fig:preference}) after the fact.

For a scenario $S$ with $K$ generated boxes, there are $K!$ possible stacking orders. For each order, we stack the boxes one by one on top of one another. The boxes are stacked horizontally centered, with some added Gaussian noise. We apply a linear momentum of $0.4\si[per-mode=symbol]{\meter\per\second}$ to the box being stacked on top and its contents. The momentum is downwards facing and deviating up to $\sigma_\alpha=\ang{13}$ from the gravity vector. This momentum simulates that a box being placed is not decelerated to $0\si[per-mode=symbol]{\meter\per\second}$ at the moment of contact with the bottom box, which might topple the stack.
We wait for the simulation to return to rest before the next box is placed.
If the stack falls over, we terminate the execution of the current order. We share additional sampling parameters such as material densities in \cref{tab:random_params}.

\begin{figure}
    \centering
    \includegraphics[width=\linewidth]{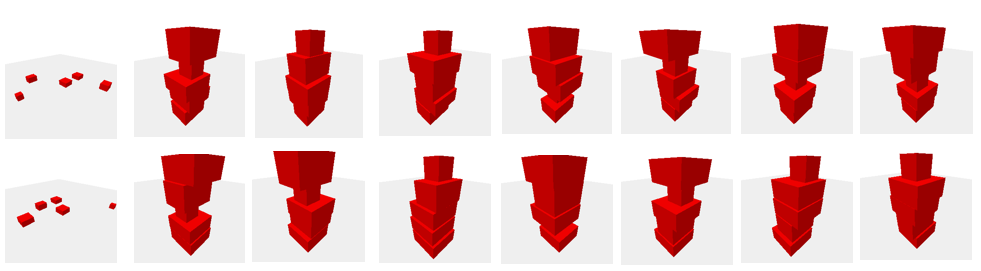}
    \caption{Examples of stable stacks generated from physics simulation.}
    \label{fig:dataset}
\end{figure}

Each stack of boxes $a_i = \tuple{b_1, \ldots, b_k}$ with $k \leq K$ can be scored under a preference $p$ by measuring its deviation from the optimal preference $a_p^*$ which is obtained by stably sorting $a_i$ under $p$ as  $a_p^* = \text{sort}_p(a_i)$. We compare the two sequences using the Levenshtein distance $\text{lev}(a_i, a_p^*)$ which we normalize by the length of the sequence $|a_i|$ as
\[
    \phi(a, p) = \frac{\text{lev}(a, \text{sort}_p(a))}{|a|}
\]
If $a_i=a_p^*$, this metric is $0$. If all positions in $a_i$ differ from those in $a_p^*$, the metric is $1$.
From our simulated set of all samples, we can now generate training examples for specific single or mixed preferences. We distinguish preferences as building on visually apparent characteristics, such as a box's size as $\hat{p}$, and characteristics requiring interaction as $\bar{p}$.
Let $i$ then be a particular order and $j$ be the step in building the full stack $i$. We generate a training sample for a joint preference $P=\set{\bar{p}_1, \ldots, \bar{p}_{\bar{N}}, \hat{p}_1, \ldots, \hat{p}_{\hat{N}}}$ by picking $i$ as 
\[
    a_i = \argmax_{a \in A_S} \frac{1}{\hat{N}} \sum_n^{\hat{N}} \phi(a, p_n),
\]
where $A_S$ is the set of all completed stable stacks generated for scenario $S$ in simulation (see Fig. \ref{fig:dataset} for examples of stable stacks). Given this selected target, we increment $j$ and check after each increment if there is another partial stack $a_{l,j}$ with the same objects that scores better under the full average score of $P$ than the current stack $a_{i,j}$. If such an $l$ does exist, we generate unstacking and stacking actions to convert $a_{i,j}$ to $a_{l,j} $, making $a_l$ the new current stack $a_i$.

For any selection $P$ we provide different prompt options and generate text for the stacking and unstacking actions in the format that we demand from the LLM. We do not include any visual observations to prevent overfitting to the appearance of the simulation.
This procedure allows us to generate the needed amount of training data in a reasonable amount of time. While simulating all $K!$ stacking orders is time-consuming, generating thousands of different executions from the simulation results is extremely efficient and flexible.

\begin{table}[t]
\centering
\caption{Collection of parameters used for simulation environments. }
\label{tab:random_params}
\begin{tabular}{rl}
\toprule
\multicolumn{1}{c}{\textbf{Parameter}} & \textbf{Range} \\ \midrule
Number of boxes    & $[3, \; 6]$ \\ 
Number of objects  & $[1, \; 7]$ \\ 
Box densities      & $\set{0.69, 0.7, 1.53}\ \si[per-mode=symbol]{\gram\per\centi\meter\cubed}$ \\ 
\multirow{2}{*}{Object Densities} & $\{0.7, 1.53, 2.7, 2.7, 7.8,$\\
                 & $  8.6, 9.0, 11.3, 19.3\}\ \si[per-mode=symbol]{\gram\per\centi\meter\cubed}$\\ 
Placement $\sigma$ & $2\si{\centi\meter}$ \\
Placement impulse angle $\sigma_\alpha$ & $\ang{13}$ \\
Placement impulse                       & $0.4\si[per-mode=symbol]{\meter\per\second}$ \\
\bottomrule
\end{tabular}
\end{table}

\section{Experimental Results}

We evaluate our approach in two stages. First, we use the simulation environment we used to generate training data to perform simulated inferences. In this environment, we can study at scale the impact of different model sizes and fine-tuning in a controlled setting. Second, we evaluate our full approach on a real robotic system with live sensor data and robotic manipulation to assess its real-world capabilities.

\subsection{Quantitative Simulation Evaluation}
\subsubsection{Simulation evaluation setup}
We investigate the impact of fine-tuning a large language model with our generated dataset for long-horizon stacking tasks. We build upon the simulation environment described in \cref{sec:gen-data} and the interpretation of preferences as competing sorting orders.
During data generation, we instantiate between 3 and 6 random boxes in a simulated scene, select a random preference, and ask the agent to stack these boxes accordingly. To ensure that the task is feasible, we score all possible stacking orders using the mean normalized Levenshtein distance and simulate the best scoring stacks to assess their feasibility. The best-scoring physically stable stack determines the target value that can be achieved in this scene. If this value is below $0.4$ then we deem the preference as too at odds with this scene and sample a new set of boxes.
The preferences we test for are reported in \cref{tab:sim_eval}. We run an equal number of scenes for each preference, 200 in total.

\subsubsection{Prompt-tuned and fine-tuned models with different model sizes}
\begin{figure}
    \centering
    \includegraphics[width=0.7\linewidth]{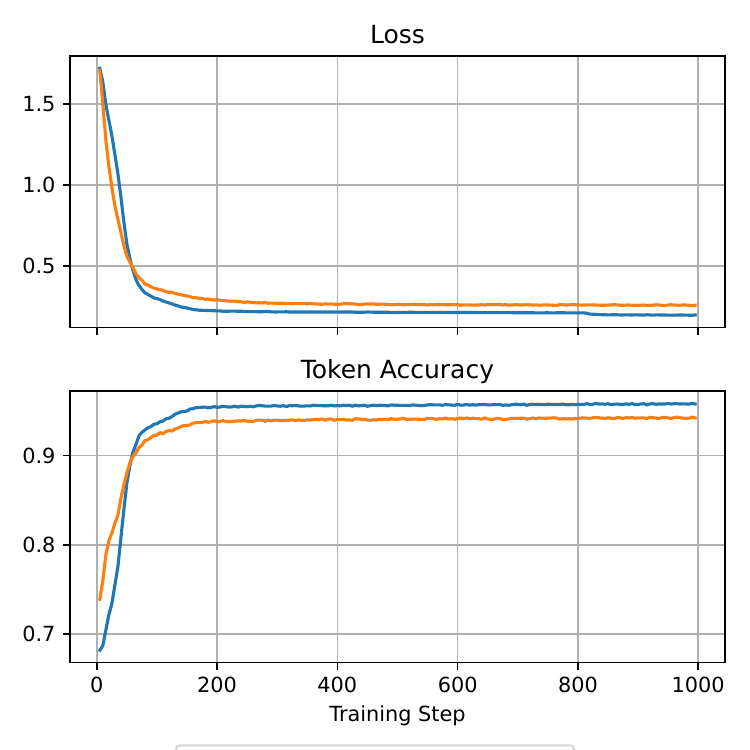}
    \caption{Training loss and token accuracy during LLM fine-tuning. Independent of parameter count, the models converge at a similar rate and to similar loss and accuracy.}
    \label{fig:loss}
\end{figure}

In our study, we evaluate two pre-trained LLMs with prompt tuning: Mistral-7B-Instruct-v0.2 (Mistral-7B) and Mistral-Small-24B-Instruct-2501 (Mistral-24B). Both models are given an example of a stacking interaction and are then told that they are now to consider a new scenario.
Additionally, we fine-tune Mistral-7B and Mistral-24B on our training data using LoRA~\cite{hu2022lora} for 1000 steps on two Nvidia A100 GPUs, which takes $~2$ hours for Mistral-7B and $~6$ hours for Mistral-24B. Note that using two Nvidia 4090 GPUs or one Nvidia 5090 GPU also gives reasonable fine-tuning time. The models converge to similar losses and similar accuracy, as can be seen in \cref{fig:loss}. We evaluate different checkpoints and note that overfitting sets in after $500$ training steps for Mistral-7B, and after $200$ steps for Mistral-24B. We report the results for these checkpoints.
Since the fine-tuned models should already be familiar with the task, we skip the prompt-tuning and directly start the task. During inference, we use semantically equivalent descriptions of the preferences that were not seen during training.

\subsubsection{Performance metrics and evaluation}
\begin{figure*}
    \centering
    \includegraphics[width=0.9\textwidth]{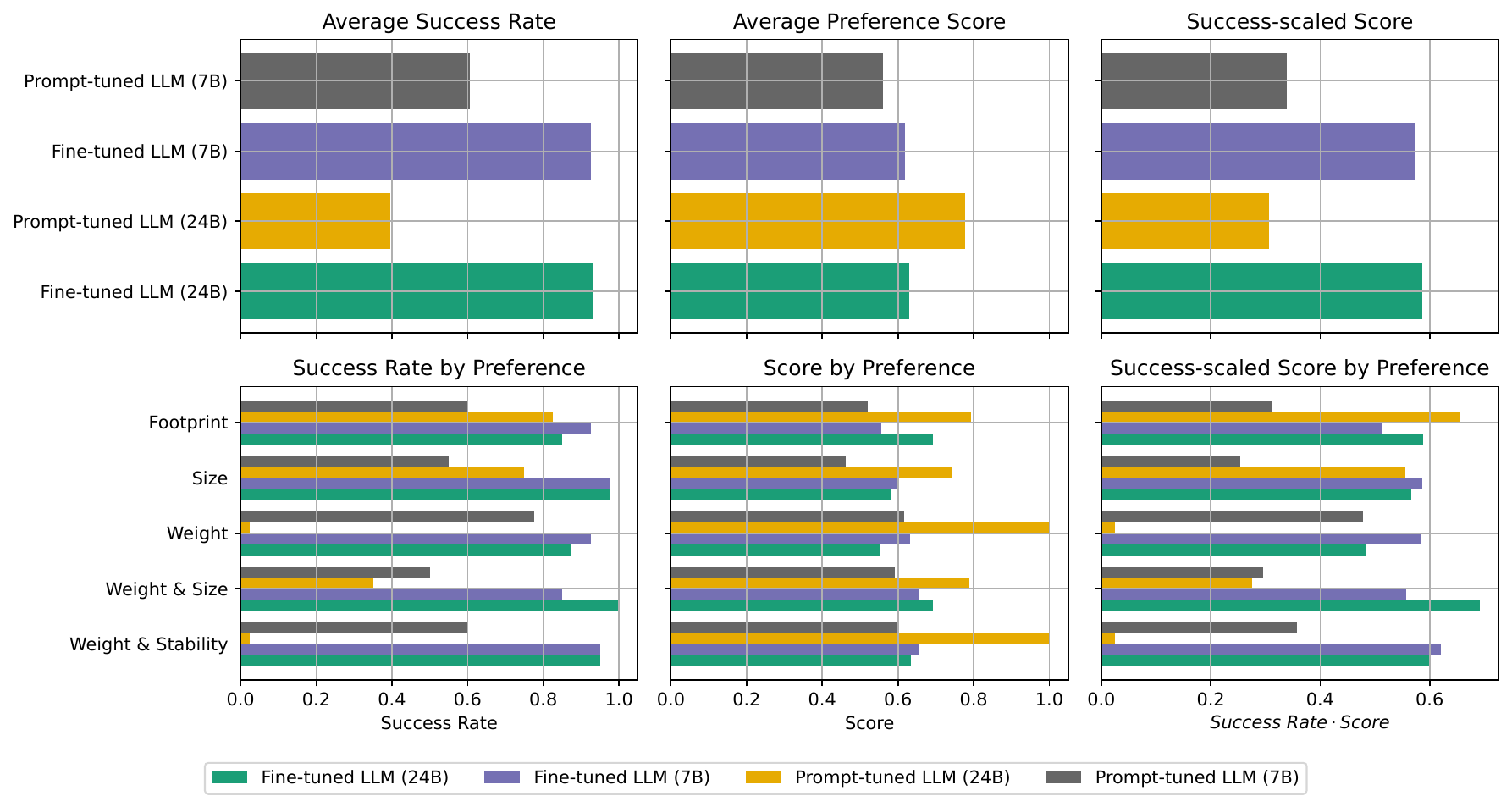}
    \caption{Visual comparison of success rate, score, and combined score of different models per preference. \textbf{Top}: Both pre-trained and fine-tuned models perform similarly well at abiding by a preference, but the pre-trained models are less able to finish the task. \textbf{Bottom}: Differentiating by preference reveals that the larger 24B model is unable to handle latent characteristics of objects. Fine-tuning remedies these problems, boosting the success rate and overall score.}
    \label{fig:pref_stats}
    \vspace{-0.3cm}
\end{figure*}

In \cref{tab:sim_eval}, we report the \textit{Success Rate}, which is only building a full stable stack, and the \textit{Preference Score} of this stack relative to the best possible score we determine beforehand. Additionally, we form the product of both to obtain an overall performance score for each model (\textit{Success-scaled Score}).
In the first row of \cref{fig:pref_stats}, we report the aggregates of these metrics across all different preferences. The results show that the prompt-tuned models demonstrate lower performance at finishing the stacking task than the fine-tuned models. When they do finish the task, the score of the produced solution is similar to that of the fine-tuned models, and in the case of Mistral-24B even $15\%$ higher.
To understand these results better, we present the success and score differentiated by task in the second row of \cref{fig:pref_stats}. The differentiation shows that Mistral-7B's performance is very stable across the different preferences, while Mistral-24B is much less able to finish the task when it is dependent on a latent characteristic which is only revealed through the interaction.
The impact of fine-tuning either model is greatest in the success rate. While Mistral-7B also improves on the preference score, fine-tuning worsens this metric for the Mistral-24B model.

We conclude that the preference-based stacking task is not trivial, especially if the preference requires information about objects that is not available from the very beginning. Our technique for generating training data proves itself successful at providing data to enable the agents to become much more successful at the task overall.

\begin{table*}[h!]

\centering
\caption{Success rate and average compliance to optimal stacking order in succeeded tasks for different preferences, across number of boxes (3--6). [Success rate / Preference score / Success-scaled score]}
\label{tab:sim_eval}
\begin{adjustbox}{max width=\textwidth}

\begin{tabular}{llcccc}
\toprule
\multirow{2}{*}{Preference} & \multirow{2}{*}{Model} & \multicolumn{4}{c}{Number of Boxes} \\
\cmidrule(lr){3-6}
 & & 3 & 4 & 5 & 6 \\
\midrule
\multirow{4}{*}{Footprint} & Prompt-tuned LLM (7B)  & 0.70 / 0.62 / 0.43 & 0.90 / 0.44 / 0.40 & 0.33 / \textbf{0.60} / 0.20 & 0.45 / 0.47 / 0.21 \\
                           & Fine-tuned LLM (7B)    & \textbf{1.00} / 0.70 / 0.70 & \textbf{0.92} / 0.55 / 0.50 & \textbf{1.00} / 0.52 / 0.52 & 0.71 / 0.40 / 0.29 \\
                           & Prompt-tuned LLM (24B) & \textbf{1.00} / \textbf{0.81} / \textbf{0.81} & 0.85 / \textbf{0.86} / \textbf{0.73} & \textbf{1.00} / 0.57 / \textbf{0.57} & 0.62 / \textbf{0.88} / \textbf{0.54} \\
                           & Fine-tuned LLM (24B)   & \textbf{1.00} / 0.80 / 0.80 & 0.75 / 0.83 / 0.62 & 0.89 / 0.53 / 0.47 & \textbf{0.76} / 0.68 / 0.52 \\
\cmidrule{2-6}
\multirow{4}{*}{Size} & Prompt-tuned LLM (7B)  & 0.44 / 0.67 / 0.30 & 0.75 / 0.38 / 0.28 & 0.71 / 0.44 / 0.31 & 0.22 / 0.42 / 0.09 \\
                      & Fine-tuned LLM (7B)    & \textbf{1.00} / \textbf{0.74} / \textbf{0.74} & \textbf{1.00} / 0.68 / 0.68 & \textbf{1.00} / 0.53 / 0.53 & \textbf{0.91} / 0.38 / 0.35 \\
                      & Prompt-tuned LLM (24B) & 0.90 / 0.70 / 0.63 & 0.83 / \textbf{0.78} / 0.65 & 0.80 / \textbf{0.70} / 0.56 & 0.54 / \textbf{0.76} / \textbf{0.41} \\
                      & Fine-tuned LLM (24B)   & \textbf{1.00} / 0.52 / 0.52 & \textbf{1.00} / 0.75 / \textbf{0.75} & \textbf{1.00} / 0.60 / \textbf{0.60} & \textbf{0.91} / 0.40 / 0.36 \\
\cmidrule{2-6}
\multirow{4}{*}{Weight} & Prompt-tuned LLM (7B)  & 0.92 / 0.72 / 0.67 & 0.75 / 0.61 / 0.46 & 0.57 / \textbf{0.65} / 0.37 & 0.75 / 0.39 / 0.29 \\
                        & Fine-tuned LLM (7B)    & \textbf{1.00} / \textbf{0.85} / \textbf{0.85} & \textbf{1.00} / 0.64 / \textbf{0.64} & 0.90 / 0.47 / \textbf{0.42} & 0.67 / 0.29 / 0.19 \\
                        & Prompt-tuned LLM (24B) & 0.00 /   -  /   -  & 0.12 / \textbf{1.00} / 0.12 & 0.00 /    - /    - & 0.00 /    - /   -  \\
                        & Fine-tuned LLM (24B)   & 0.91 / 0.57 / 0.52 & 0.71 / 0.60 / 0.43 & \textbf{1.00} / 0.42 / \textbf{0.42} & \textbf{0.83} / \textbf{0.65} / \textbf{0.54} \\
\cmidrule{2-6}
\multirow{4}{*}{Weight \& Size} & Prompt-tuned LLM (7B)  & 0.67 / 0.62 / 0.42 & 0.40 / 0.54 / 0.22 & 0.56 / 0.57 / 0.32 & 0.47 / 0.62 / 0.29 \\
                                & Fine-tuned LLM (7B)    & \textbf{1.00} / \textbf{1.00} / \textbf{1.00} & 1.00 / \textbf{0.78} / \textbf{0.78} & \textbf{1.00} / 0.46 / 0.46 & 0.60 / 0.47 / 0.28 \\
                                & Prompt-tuned LLM (24B) & 0.18 / 0.33 / 0.06 & 0.45 / \textbf{0.78} / 0.35 & 0.44 / \textbf{0.94} / 0.42 & 0.33 / \textbf{0.92} / 0.31 \\
                                & Fine-tuned LLM (24B)  & \textbf{1.00} / 0.68 / 0.68 & 1.00 / 0.58 / 0.58 & \textbf{1.00} / 0.81 / \textbf{0.81} & \textbf{1.00} / 0.73 / \textbf{0.73} \\
\cmidrule{2-6}
\multirow{4}{*}{Weight \& Stability} & Prompt-tuned LLM (7B)  & 0.50 / \textbf{0.88} / 0.44 & 0.80 / 0.56 / 0.45 & 0.62 / 0.47 / 0.29 & 0.44 / 0.65 / 0.29 \\
                                     & Fine-tuned LLM (7B)    & \textbf{1.00} / 0.76 / \textbf{0.76} & \textbf{1.00} / 0.71 / 0.71 & 1.00 / 0.59 / \textbf{0.59} & 0.67 / 0.45 / 0.30 \\
                                     & Prompt-tuned LLM (24B) & 0.00 /    - /    - & 0.00 /    - /    - & 0.00 /    - /    - & 0.10 / \textbf{1.00} / 0.10 \\
                                     & Fine-tuned LLM (24B)  & 0.90 / 0.61 / 0.55 & \textbf{1.00} / 0.76 / \textbf{0.76} & 1.00 / 0.53 / 0.53 & \textbf{0.88} / 0.64 / \textbf{0.56} \\
\bottomrule
\end{tabular}
\end{adjustbox}
\end{table*}

\subsubsection{Scalability to different numbers of boxes}
Increasing the number of boxes brings negligible extra costs for an LLM to plan the stacking task. However, in general, we found a tendency of downgraded performance to different extents as the number of boxes increases from \cref{tab:sim_eval}. In contrast, instead of using LLM to generate the stacking sequence, rule-based methods require manually defining the corresponding stacking sequence for each possible stacking status involving stacking and unstacking. In particular, when the existing stack needs to be adjusted, it would require quite a few engineering efforts even for as few as three boxes, and it becomes difficult to scale when the number of boxes to stack increases. 

\subsection{Real Robotic Evaluation}

\begin{figure}
    \centering
    \includegraphics[width=\linewidth]{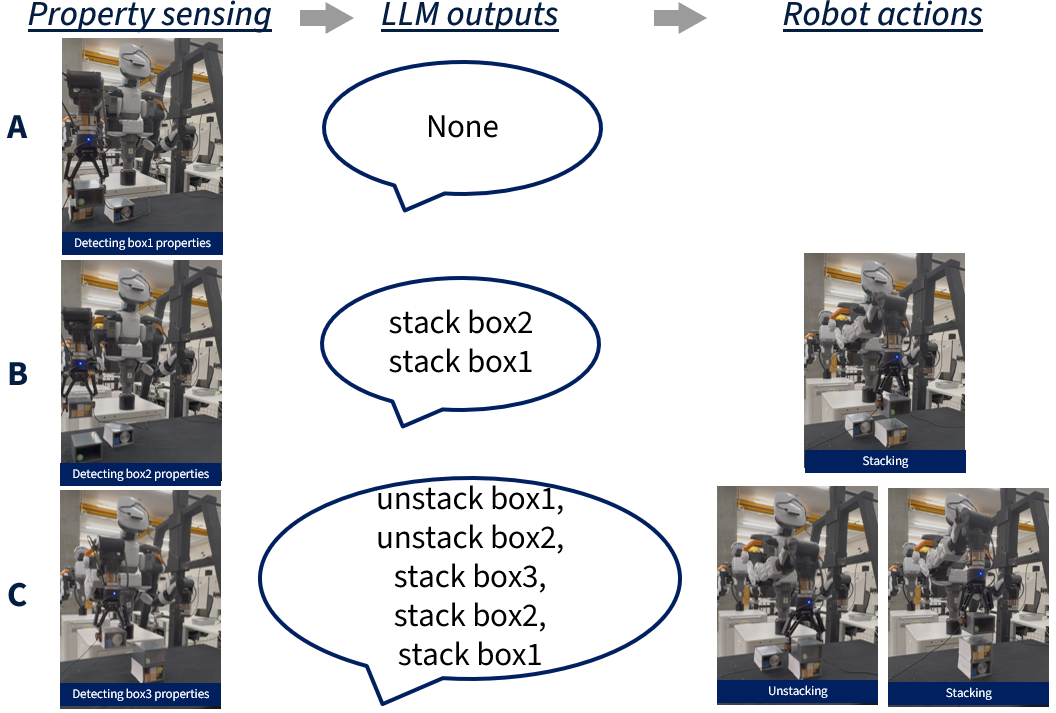}
    \caption{The NEXTAGE robot performing a long-horizon stacking task of three boxes with various contents in an online manner. \textbf{Left:} The NEXTAGE robot detecting weight and stability for each box. \textbf{Middle:} LLM generating corresponding stacking sequence on-the-fly after each detection. \textbf{Right:} The NEXTAGE robot executing stacking or unstacking according to LLM outputs.}
    \label{fig:result}
    \vspace{-0.3cm}
\end{figure}

\subsubsection{Robot platform}
We deployed our proposed framework on a Kawada NEXTAGE humanoid robot equipped with a Robotiq 140 gripper, wrist-mounted force-torque sensor, and two piezo microphones on each finger of the gripper, as shown in Fig. \ref{fig:teaser}. 

\subsubsection{Object pose estimation}
The 6D pose of each box is estimated by FoundationPose \cite{wen2024foundationpose} and Segment Anything 2 (SAM2) \cite{ravi2024sam2}. FoundationPose estimates 6D pose from RGB, depth, and masked images, where the masked images are obtained from SAM2. The object poses are then converted to the world frame for the motion controllers. We use an Intel Realsense D435 camera installed on the head of the humanoid robot facing downwards.

\subsubsection{Motion controllers}
After obtaining estimated object poses in the world frame, we use OpTaS (OPtimization-based TAsk Specification library) \cite{mower2023optas} to solve the inverse kinematics to control the gripper to reach each object.

\subsubsection{LLM inference}
We use a separate desktop with two Nvidia 4090 GPUs to perform LLM inference. Each inference takes less than five seconds. The desktop receives a force and audio signal from the robot computer and sends back the stacking sequence. 

\subsubsection{Offline stacking}
The offline planning of box stacking is straightforward. The robot lifts up and tilts each box in turn to get the weight and stability of all boxes. All information is encoded, sent to Mistral-24B, and the LLM generates the best stacking order for the motion controller to execute. 

\subsubsection{Online stacking}

Here, we show the robot performing successful stacking according to weight and stability using our proposed framework. The long-horizon stacking task is planned on the fly by Mistral-24B. We have three boxes with the same visual appearance but unknown hidden properties due to various contents inside (Fig. \ref{fig:teaser}). As shown in Fig. \ref{fig:result}, the robot first detects the weight and stability of box 1. At this stage, since no other boxes have been manipulated and detected, the LLM decides to wait for the properties of another box before reaching a decision. After getting the properties of box 2 and determining that box 2 needs to be put under box 1, the LLM generates the stacking sequence \textit{"stack box2, stack box1"}. The robot performs the current best sequence for the boxes with known hidden properties. After detecting box 3's properties, LLM figures out that box 3 needs to be at the bottom and the existing stack needs to be adjusted, so it generates a new sequence to guide the robot to a series of unstacking and stacking to reach the ideal stack in the end. See the robot in action in the accompanying video.

\section{Conclusion}
In this work, we presented a framework for using multimodal information in combination with large language models for high-level online planning for long-horizon robotic stacking tasks. In particular, we addressed the stacking of objects with latent characteristics that can only be accessed through interactions. When taking multiple conflicting human stacking preferences into account, we were able to generate a sufficient amount of diverse data samples in physics simulation using our proposed dataset generation techniques for fine-tuning the LLM. Through quantitative simulation evaluations, we showcased improved reasoning capabilities of the fine-tuned LLM over a combination of stacking preferences involving physics properties such as weight and stability. Moreover, we successfully deployed our framework on a robotic manipulator in the real world.

For future work, incorporating a broader range of user preferences and constraints would enable the model to handle more diverse stacking scenarios, extending its applicability to everyday settings, such as arranging box stacks to fit within the height limits of a cupboard. Besides, we would like to investigate the long-horizon stacking tasks in a dual-arm setting to handle larger objects and extend the working range.

\bibliographystyle{IEEEtran}

\begin{thebibliography}{10}
\providecommand{\url}[1]{#1}
\csname url@samestyle\endcsname
\providecommand{\newblock}{\relax}
\providecommand{\bibinfo}[2]{#2}
\providecommand{\BIBentrySTDinterwordspacing}{\spaceskip=0pt\relax}
\providecommand{\BIBentryALTinterwordstretchfactor}{4}
\providecommand{\BIBentryALTinterwordspacing}{\spaceskip=\fontdimen2\font plus
\BIBentryALTinterwordstretchfactor\fontdimen3\font minus \fontdimen4\font\relax}
\providecommand{\BIBforeignlanguage}[2]{{%
\expandafter\ifx\csname l@#1\endcsname\relax
\typeout{** WARNING: IEEEtran.bst: No hyphenation pattern has been}%
\typeout{** loaded for the language `#1'. Using the pattern for}%
\typeout{** the default language instead.}%
\else
\language=\csname l@#1\endcsname
\fi
#2}}
\providecommand{\BIBdecl}{\relax}
\BIBdecl

\bibitem{fikes1971strips}
R.~E. Fikes and N.~J. Nilsson, ``Strips: A new approach to the application of theorem proving to problem solving,'' \emph{Artificial intelligence}, vol.~2, no. 3-4, pp. 189--208, 1971.

\bibitem{jiang2019task}
Y.-q. Jiang, S.-q. Zhang, P.~Khandelwal, and P.~Stone, ``Task planning in robotics: an empirical comparison of pddl-and asp-based systems,'' \emph{Frontiers of Information Technology \& Electronic Engineering}, vol.~20, no.~3, pp. 363--373, 2019.

\bibitem{toussaint2015logic}
M.~Toussaint, ``Logic-geometric programming: An optimization-based approach to combined task and motion planning.'' 2015.

\bibitem{huang2022towards}
J.~Huang and K.~C.-C. Chang, ``Towards reasoning in large language models: A survey,'' \emph{arXiv preprint arXiv:2212.10403}, 2022.

\bibitem{zawalski2024robotic}
M.~Zawalski, W.~Chen, K.~Pertsch, O.~Mees, C.~Finn, and S.~Levine, ``Robotic control via embodied chain-of-thought reasoning,'' in \emph{Conference on Robot Learning}, 2024.

\bibitem{honerkamp2024language}
D.~Honerkamp, M.~B{\"u}chner, F.~Despinoy, T.~Welschehold, and A.~Valada, ``Language-grounded dynamic scene graphs for interactive object search with mobile manipulation,'' \emph{IEEE Robotics and Automation Letters}, 2024.

\bibitem{chisari2025robotic}
E.~Chisari, J.~O. von Hartz, F.~Despinoy, and A.~Valada, ``Robotic task ambiguity resolution via natural language interaction,'' \emph{Int.~Conf.~on Intelligent Robots and Systems}, 2025.

\bibitem{shi2025hi}
L.~X. Shi, B.~Ichter, M.~Equi, L.~Ke, K.~Pertsch, Q.~Vuong, J.~Tanner, A.~Walling, H.~Wang, N.~Fusai \emph{et~al.}, ``Hi robot: Open-ended instruction following with hierarchical vision-language-action models,'' \emph{arXiv preprint arXiv:2502.19417}, 2025.

\bibitem{kim2024openvla}
M.~J. Kim, K.~Pertsch, S.~Karamcheti, T.~Xiao, A.~Balakrishna, S.~Nair, R.~Rafailov, E.~Foster, G.~Lam, P.~Sanketi \emph{et~al.}, ``Openvla: An open-source vision-language-action model,'' \emph{arXiv preprint arXiv:2406.09246}, 2024.

\bibitem{team2025gemini}
G.~R. Team, S.~Abeyruwan, J.~Ainslie, J.-B. Alayrac, M.~G. Arenas, T.~Armstrong, A.~Balakrishna, R.~Baruch, M.~Bauza, M.~Blokzijl \emph{et~al.}, ``Gemini robotics: Bringing ai into the physical world,'' \emph{arXiv preprint arXiv:2503.20020}, 2025.

\bibitem{black2410pi0}
K.~Black, N.~Brown, D.~Driess, A.~Esmail, M.~Equi, C.~Finn, N.~Fusai, L.~Groom, K.~Hausman, B.~Ichter \emph{et~al.}, ``$\pi$0: A vision-language-action flow model for general robot control, 2024,'' \emph{arXiv preprint arXiv:2410.24164}.

\bibitem{nazarczuk2024closed}
M.~Nazarczuk, J.~K. Behrens, K.~Stepanova, M.~Hoffmann, and K.~Mikolajczyk, ``Closed loop interactive embodied reasoning for robot manipulation,'' \emph{arXiv preprint arXiv:2404.15194}, 2024.

\bibitem{cannizzaro2023towards}
R.~Cannizzaro, J.~Routley, and L.~Kunze, ``Towards a causal probabilistic framework for prediction, action-selection \& explanations for robot block-stacking tasks,'' \emph{arXiv preprint arXiv:2308.06203}, 2023.

\bibitem{cannizzaro2024causal}
R.~Cannizzaro, M.~Groom, J.~Routley, R.~O. Ness, and L.~Kunze, ``A causal bayesian network and probabilistic programming based reasoning framework for robot manipulation under uncertainty,'' \emph{arXiv preprint arXiv:2403.14488}, 2024.

\bibitem{lee2021beyond}
A.~X. Lee, C.~M. Devin, Y.~Zhou, T.~Lampe, K.~Bousmalis, J.~T. Springenberg, A.~Byravan, A.~Abdolmaleki, N.~Gileadi, D.~Khosid \emph{et~al.}, ``Beyond pick-and-place: Tackling robotic stacking of diverse shapes,'' in \emph{5th Annual Conference on Robot Learning}, 2021.

\bibitem{lampe2024mastering}
T.~Lampe, A.~Abdolmaleki, S.~Bechtle, S.~H. Huang, J.~T. Springenberg, M.~Bloesch, O.~Groth, R.~Hafner, T.~Hertweck, M.~Neunert \emph{et~al.}, ``Mastering stacking of diverse shapes with large-scale iterative reinforcement learning on real robots,'' in \emph{IEEE Int.~Conf.~on Robotics and Automation}, 2024, pp. 7772--7779.

\bibitem{argus2025cvla}
M.~Argus, J.~Bratulic, H.~Masnavi, M.~Velikanov, N.~Heppert, A.~Valada, and T.~Brox, ``cvla: Towards efficient camera-space vlas,'' \emph{arXiv preprint arXiv:2507.02190}, 2025.

\bibitem{chiang2018design}
Y.-C. Chiang, H.~Bier, and S.~Mostafavi, ``Design to robotic assembly: An exploration in stacking,'' \emph{Frontiers in Digital Humanities}, vol.~5, p.~23, 2018.

\bibitem{liu2023stability}
Y.~Liu and N.~Napp, ``Stability-based sequence planning for robotic dry-stacking of natural stones,'' \emph{IEEE Robotics and Automation Letters}, vol.~8, no.~9, pp. 5894--5901, 2023.

\bibitem{furrer2017autonomous}
F.~Furrer, M.~Wermelinger, H.~Yoshida, F.~Gramazio, M.~Kohler, R.~Siegwart, and M.~Hutter, ``Autonomous robotic stone stacking with online next best object target pose planning,'' in \emph{IEEE Int.~Conf.~on Robotics and Automation}, 2017, pp. 2350--2356.

\bibitem{hu2022lora}
E.~J. Hu, Y.~Shen, P.~Wallis, Z.~Allen-Zhu, Y.~Li, S.~Wang, L.~Wang, W.~Chen \emph{et~al.}, ``Lora: Low-rank adaptation of large language models.'' \emph{ICLR}, vol.~1, no.~2, p.~3, 2022.

\bibitem{wen2024foundationpose}
B.~Wen, W.~Yang, J.~Kautz, and S.~Birchfield, ``Foundationpose: Unified 6d pose estimation and tracking of novel objects,'' in \emph{IEEE/CVF Conf.~on Computer Vision and Pattern Recognition}, 2024.

\bibitem{ravi2024sam2}
N.~Ravi, V.~Gabeur, Y.-T. Hu, R.~Hu, C.~Ryali, T.~Ma, H.~Khedr, R.~R{\"a}dle, C.~Rolland, L.~Gustafson, E.~Mintun, J.~Pan, K.~V. Alwala, N.~Carion, C.-Y. Wu, R.~Girshick, P.~Doll{\'a}r, and C.~Feichtenhofer, ``Sam 2: Segment anything in images and videos,'' \emph{arXiv preprint arXiv:2408.00714}, 2024.

\bibitem{mower2023optas}
C.~E. Mower, J.~Moura, N.~Z. Behabadi, S.~Vijayakumar, T.~Vercauteren, and C.~Bergeles, ``Optas: An optimization-based task specification library for trajectory optimization and model predictive control,'' \emph{arXiv preprint arXiv:2301.13512}, 2023.

\end{thebibliography}

\end{document}